\documentclass[10pt,twocolumn,letterpaper]{article}

\usepackage[pagenumbers]{cvpr} 

\usepackage{algorithm}
\usepackage{algorithmic}
\usepackage{caption}
\usepackage{mathcomp}
\usepackage{multirow}
\usepackage[table]{xcolor}
\newcommand{\scc}[3]{\cellcolor{#1!#2!white}{#3}}
\usepackage{soul}

\def\methodAbbr{COG}
\def\methodName{Confidence-aware Optimal Geometric Correspondence}

\definecolor{cvprblue}{rgb}{0.21,0.49,0.74}
\usepackage[pagebackref,breaklinks,colorlinks,allcolors=cvprblue]{hyperref}

\def\paperID{14085} 
\def\confName{CVPR}
\def\confYear{2026}

\title{\methodAbbr: \methodName\ for Unsupervised Single-reference Novel Object Pose Estimation}

\author{
Yuchen Che$^{1}$ \quad
Jingtu Wu$^{1}$ \quad
Hao Zheng$^{2}$ \quad
Asako Kanezaki$^{1,2,3}$\\
$^{1}$Institute of Science Tokyo \quad $^{2}$RIKEN \quad $^{3}$Tohoku University \\
{\tt\small
    \{che.y.99ea@m, wu.j.baa4@m, kanezaki@comp\}.isct.ac.jp,
    hao.zheng@riken.jp
}\\
}

\begin{document}

\twocolumn[{%
\renewcommand\twocolumn[1][]{#1}%
\maketitle
\vspace{-1.25em}
\includegraphics[width=0.996\linewidth]{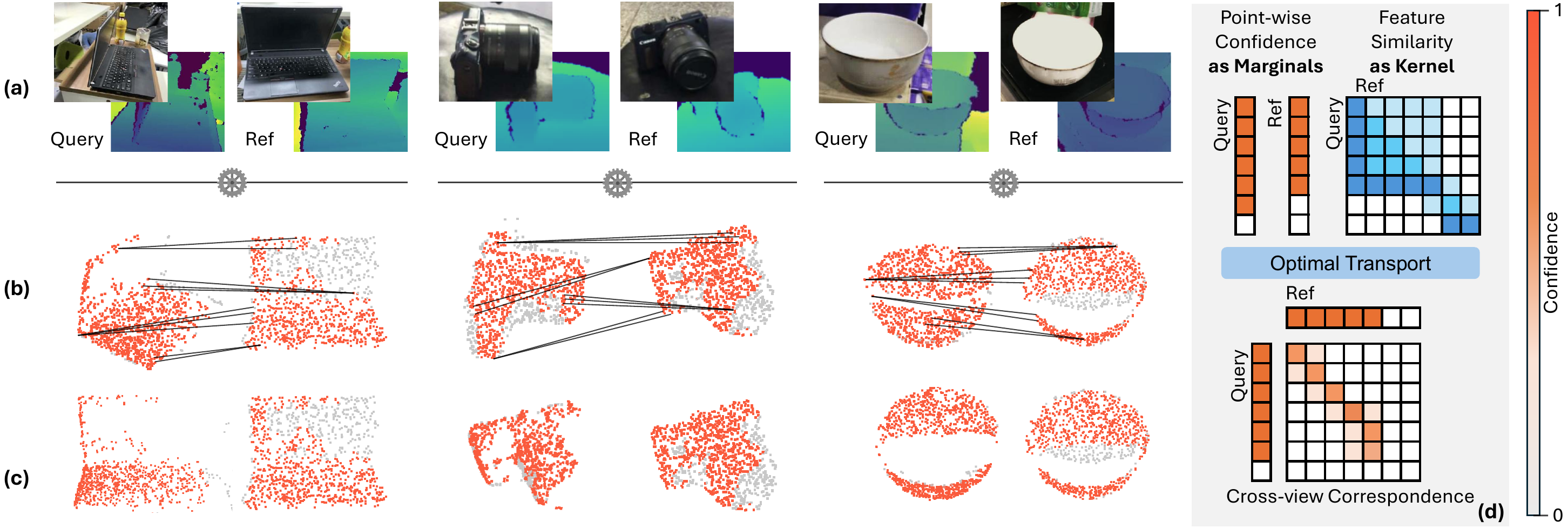}
\vspace{-0.25em}
\captionof{figure}{Given a novel object's query and reference RGB-D images (a), \methodAbbr\ outputs point-wise confidence and cross-view soft correspondence (b), to estimate the relative pose between query and reference (c). To achieve this, we formulate correspondence finding as an optimal transport problem, with each point's confidence as target marginals, and the point features' similarity as an affinity kernel (d).
\vspace{1em}
}
\label{fig:teaser}
}]

\begin{abstract}
Estimating the 6DoF pose of a novel object with a single reference view is challenging due to occlusions, viewpoint changes, and outliers. 
A core difficulty lies in finding robust cross-view correspondences, as existing methods often rely on discrete one-to-one matching that is non-differentiable and tends to collapse onto sparse keypoints. 
We propose \methodName~(\methodAbbr), an unsupervised framework that formulates correspondence estimation as a confidence-aware optimal transport problem. 
\methodAbbr\ produces balanced soft correspondences by predicting point-wise confidences and injecting them as optimal transport marginals, suppressing non-overlapping regions. 
Semantic priors from vision foundation models further regularize the correspondences, leading to stable pose estimation. 
This design integrates confidence into the correspondence finding and pose estimation pipeline, enabling unsupervised learning.
Experiments show unsupervised \methodAbbr\ achieves comparable performance to supervised methods, and supervised \methodAbbr\ outperforms them.
Codes: \small{\url{https://github.com/YC-Che/COG}}
\end{abstract}
    
\section{Introduction}\label{sec:intro}
Object pose estimation, which aims to recover an object's 6DoF pose (rotation and translation) from RGB-D images, is a fundamental task for robotics~\cite{robotic-assembly, lanpose, deep-grasping}, augmented reality~\cite{ar-survey, multi-state-ar}, and 3D scene understanding~\cite{omni6dpose, holistic-scene}. 
To enable general real-world deployment, improving the generalization capability of object pose estimation has long been a central goal. 
Early instance-level methods~\cite{6d-diff, gdr, ffb6d, dense-fusion} assume access to CAD models or training objects identical to those at test time. 
This assumption was relaxed in category-level pose estimation~\cite{NOCS, gs-pose, Shapo, op-align}, where models learn to estimate the poses of unseen instances within a few predefined categories.

Recent research has advanced toward novel object pose estimation~\cite{onepose, megapose, sam6d, onepose++, genflow, vfm}, which targets poses of arbitrary objects with dataset-agnostic generalization. 
However, these methods often depend on CAD models or multiple reference views, which limits scalability in practice. 
A more challenging setting considers only a single reference image~\cite{UnoPose, novel, pope, nope} (see Fig.~\ref{fig:teaser}), where large viewpoint changes and partial observations require the network to jointly infer valid overlapping regions and object pose, making the problem particularly ill-posed. 
A key to solving this task is to establish reliable cross-view correspondences, since the pose can be recovered by aligning geometric structures between query and reference views. 
Yet, most existing approaches~\cite{UnoPose, sam6d} construct correspondences via discrete one-to-one assignments (\eg $\operatorname{argmax}$), which tend to collapse onto a few dominant keypoints, leaving many of the points unused. 
Moreover, such non-differentiable assignments breaks differentiability and prevents the model from being trained in an unsupervised manner.

We introduce \methodName~(\methodAbbr), an unsupervised framework for novel object pose estimation from a single reference image. 
\methodAbbr\ addresses the above issues by formulating soft correspondence as an optimal transport (OT) problem, where point-wise confidences are predicted beforehand and explicitly incorporated as target marginals of the transport plan. 
Compared to OT-based methods~\cite{dustbin-ot, rpmnet, robustot} with uniform marginals and only apply confidence post hoc, This formulation yields globally balanced correspondences that naturally suppresses outliers and non-overlapping regions. 
Given these correspondences, corresponding points are generated via convex combinations, and a weighted SVD solver~\cite{Umeyama} is used to recover the pose transformation. 
The entire process forms an end-to-end correspondence finding and pose estimation pipeline, enabling unsupervised optimization of both correspondence and confidence. 
To mitigate ambiguity in purely geometric matching, we integrate semantic priors denoised from vision foundation models such as DINO~\cite{dinov1, dinov2}, which softly encourage correspondences between semantically consistent parts. 
Furthermore, for unsupervised confidence learning, Gaussian RBF-style kernels of geometric and semantic consistency are used to generate pseudo confidence labels, guiding the network to down weight uncertain points without discarding them entirely and to emphasize reliable regions with high confidence. 
With these designs, \methodAbbr\ naturally extends to the unsupervised setting, where neither ground-truth confidence nor pose supervision is available. 
Experimental results demonstrate that \methodAbbr\ achieves performance comparable to leading supervised approaches, while the supervised variant of \methodAbbr\ further outperforms them. 

Our contributions are summarized as follows:
\begin{enumerate}
    \item We formulate correspondence finding as an OT problem with confidence as marginals. Compared to OT with uniform marginals, our formulation yields balanced correspondences by suppressing non-overlapping points. 
    \item We propose an end-to-end pipeline that jointly learns object pose and point validity confidence without supervision from CAD models, poses, or overlap scores. 
    \item Unsupervised \methodAbbr\ achieves performance competitive with state-of-the-art supervised methods, and its supervised variant further outperforms them.
\end{enumerate}
\section{Related Work}

\paragraph{Instance-level and Category-level Object Pose Estimation.}
Traditional object pose estimation methods, under both instance-level~\cite{surfemb, 6d-diff, gdr, ffb6d, dense-fusion} and category-level~\cite{NOCS, Shapo, gs-pose, wild6d, op-align, se3-rigid} settings, aim to recover the 6DoF poses of known objects or objects within predefined categories. 
A common paradigm is to directly regress object poses~\cite{hs-pose, gpv-pose, fs-net} using deep backbones.
Alternatively, correspondence-based approaches establish matches either between CAD models and point clouds~\cite{CenterSnap, Shapo, ShapePrior}, or between query images and rendered CAD views~\cite{AAE, self6d, gs-pose}, followed by transformation estimation via PnP~\cite{epnp} or Umeyama~\cite{Umeyama}. 
Recently, unsupervised or weakly supervised frameworks have also gained attention. 
Equi-Pose~\cite{se3-rigid} jointly learns canonical reconstruction and pose using $SE(3)$-equivariant backbones~\cite{epn, e2pn}, while OP-Align~\cite{op-align} extends this paradigm to articulated objects. 
Zero-shot Pose~\cite{zero-shot} aligns semantic features across image sequences to establish correspondences without explicit pose labels. 
However, these settings inherently limit generalization, as they rely on prior knowledge of specific object instances or categories, restricting their ability to handle novel objects in open-world scenarios.

\paragraph{Novel Object Pose Estimation.}
Novel object pose estimation lifts the instance-level and category-level restriction by targeting arbitrary objects beyond predefined categories. 
For such dataset-agnostic generalization, many approaches~\cite{freeze, vfm, UnoPose, sam6d} exploit vision foundation models~\cite{dinov1, dinov2, vitnlp} to extract semantic priors for correspondence learning. 
Further progress~\cite{sam6d, vfm, megapose, gigapose, genflow, onepose, onepose++} leverages additional references such as CAD models or multi-view images to construct dense correspondences. 
For example, SAM-6D~\cite{sam6d} builds on CNOS~\cite{cnos} by using SAM~\cite{sam,fastsam} for segmentation and combining DINO~\cite{dinov2} features with geometric cues, 
while OnePose~\cite{onepose, onepose++} reconstructs object point clouds via Structure-from-Motion and aligns them to query views. 
MegaPose~\cite{megapose} retrieves the closest CAD rendering to the query for subsequent refinement, and GenFlow~\cite{genflow} iteratively refines poses by estimating optical flow between rendered and observed views. 
More recently, several methods attempt to reduce reference requirements to a single image~\cite{pope, UnoPose, novel, nope}.
POPE~\cite{pope} and NOPE~\cite{nope} introduce feedforward frameworks with RGB inputs,
While SinRef-6D~\cite{novel} models point-wise alignment using state-space dynamics with RGB-D inputs, and the latest UnoPose~\cite{UnoPose} constructs an $SE(3)$-invariant canonical frame for consistent object representation.
Yet, most existing approaches construct correspondences via discrete one-to-one assignments, which tend to collapse onto a few dominant keypoints, and breaks differentiability, preventing models from being trained in an unsupervised manner.

\paragraph{Point Cloud Registration.}
As a commonly used formulation in object pose estimation, correspondence finding and pose estimation can also be viewed from the perspective of point cloud registration~\cite{3dregnet, IDAM, filterreg, fcgf, dgr, rpmnet, ppf, fpfh}, which seeks to estimate the optimal rigid transformation that aligns two partially overlapping point sets. 
A common pipeline extracts discriminative local descriptors to establish correspondences, followed by robust solvers such as RANSAC~\cite{ransac} and ICP~\cite{besl1992method, ICP}. 
FCGF~\cite{fcgf} and DGR~\cite{dgr} enhance this process by predicting pairwise inlier probabilities to filter unreliable matches. 
More recent methods~\cite{rpmnet, robustot, udpreg, confidencecalib, dustbin-ot} formulate registration as an optimal transport (OT) problem, treating correspondences as continuous probability distributions. RPM-Net~\cite{rpmnet} introduces a differentiable Sinkhorn~\cite{sinkhorn} layer to compute correspondences under uniform marginals.
Meanwhile, confidence which represents the validity of correspondences has also been explored---either as post-hoc calibration of pairwise matches~\cite{confidencecalib}, as distribution-level weights~\cite{udpreg}, or as the results of a global thresholding~\cite{dustbin-ot}. 
These OT-based methods typically assign such weights after the correspondences have been established, making the correspondences still unbalanced. And more importantly, the resulting confidence cannot be optimized jointly with the correspondences in an end-to-end manner. 
In contrast, we propose a learned point-wise confidence that directly serves as the target marginal in OT, yielding globally balanced transport plans and enable end-to-end unsupervised learning.

\section{Methodology}
\subsection{Problem Formulation}
\begin{figure}[t]
    \centering
    \includegraphics[width=1\linewidth]{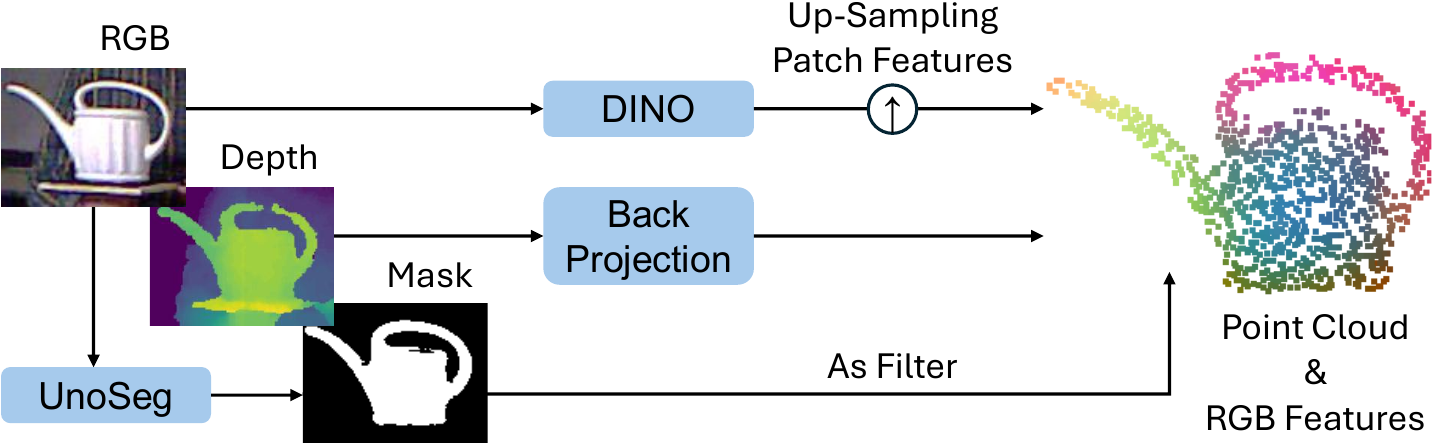}
    \caption{
    Pre-processing pipeline of \methodAbbr. Given RGB-D inputs, an object is segmented, depth map is back-projected into point clouds, and per-point RGB features are extracted from DINO to form feature augmented inputs.}
    \label{fig:pre}
\end{figure}

Given two RGB-D images of the same novel object captured from different viewpoints, one as the query and the other as the reference, our goal is to estimate the relative rigid transformation between them. 
As illustrated in Fig.~\ref{fig:pre}, we first employ a CNOS~\cite{cnos}-like segmentation model, UnoSeg~\cite{UnoPose}, to obtain object masks from the RGB images. 
The masked depth maps are then back-projected into 3D space to generate point clouds. 
In addition to geometric coordinates, we extract per-pixel RGB features using DINO~\cite{dinov2} (patch features with up-sampling), which serve as RGB descriptors. 
We denote the query and reference point clouds as $\mathbf{P} \in \mathbb{R}^{n \times 3}$ and $\mathbf{Q} \in \mathbb{R}^{n \times 3}$, and their associated RGB features as $\mathbf{F}_{p} \in \mathbb{R}^{n \times d_{\mathrm{f}}}$ and $\mathbf{F}_{q} \in \mathbb{R}^{n \times d_{\mathrm{f}}}$, 
where $n$ is the number of points and $d_{\mathrm{f}}$ is the dimension of the DINO features.
Our objective is to find a rotation $\mathbf{R}_{pq} \in SO(3)$ and a translation $\mathbf{t}_{pq} \in \mathbb{R}^3$ such that
\begin{equation}
    \mathbf{Q}^{\mathrm{val}} = \mathbf{P}^{\mathrm{val}} \mathbf{R}_{pq} + \mathbf{1} \mathbf{t}_{pq}^{\top},
\end{equation}
where $\mathbf{P}^{\mathrm{val}} \subset \mathbf{P}$ and $\mathbf{Q}^{\mathrm{val}} \subset \mathbf{Q}$ denote the valid overlapping regions of the object that can be matched across views. 
Since the observed point clouds $\mathbf{P}$ and $\mathbf{Q}$ are often partial due to occlusions and viewpoint differences, this task inherently requires reasoning about both the relative pose and the validity of each point. 
For clarity, throughout this paper the subscripts $pq$ and $qp$ denote operations in the query-to-reference and reference-to-query directions, respectively. 

\subsection{\methodAbbr}

\begin{figure*}
    \centering
    \includegraphics[width=1\linewidth]{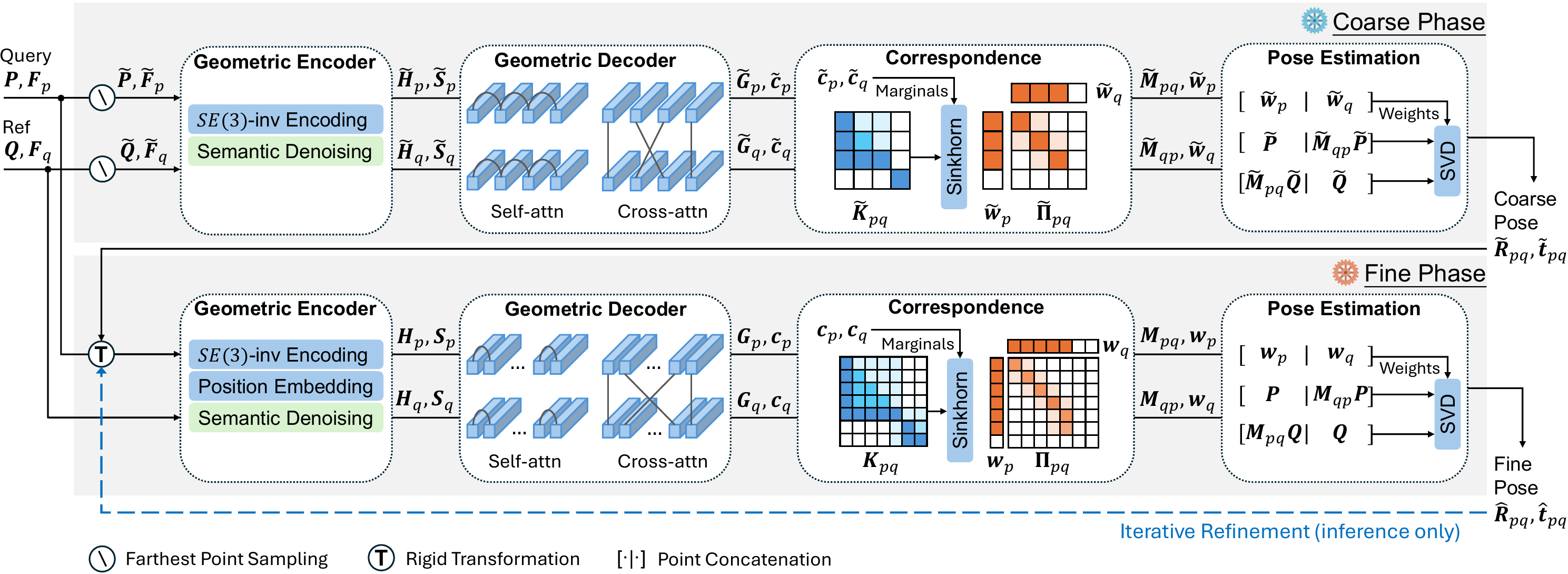}
    \caption{
    Overview of the \methodAbbr\ framework. 
    The pipeline consists of coarse and fine phases, each using a geometric transformer to predict point-wise confidences and features. 
    A Sinkhorn-based OT module computes soft correspondences, and a weighted SVD solver estimates the rigid transformation. 
    The coarse pose is further refined in the fine phase using position embeddings for precise alignment.}
    \label{fig:model}
\end{figure*}

\subsubsection{Model Overview}
As illustrated in Fig.~\ref{fig:model}, \methodAbbr\ adopts a coarse-to-fine architecture built upon a geometric transformer~\cite{geoTransformer}. 
In the coarse phase, we use farthest point sampling~\cite{fps} to obtain sparse point clouds and their DINO features as inputs (denoted with $\Tilde{\cdot}$), while in the fine phase, the full point clouds and DINO features are used for refinement. 
For clarity, we use the fine phase notation throughout the following sections; the coarse phase shares the same formulation with all terms marked by $\Tilde{\cdot}$.

Specifically, the geometric transformer encoder takes the query and reference point sets $\mathbf{P}$ and $\mathbf{Q}$ as inputs to the $SE(3)$-invariant feature encoding module (and to the position embedding module in the fine phase), producing geometric hidden features $\mathbf{H}_p$ and $\mathbf{H}_q$. 
Meanwhile, DINO features are processed through a semantic denoising module to obtain semantic embeddings $\mathbf{S}_p \in \mathbb{R}^{n \times d_{\mathrm{s}}}$ and $\mathbf{S}_q \in \mathbb{R}^{n \times d_{\mathrm{s}}}$. 
The geometric decoder then applies alternating self-attention and cross-attention layers to compute point-wise geometric features $\mathbf{G}_p \in \mathbb{R}^{n \times d_g}$ and $\mathbf{G}_q \in \mathbb{R}^{n \times d_g}$. 
Then a lightweight MLP confidence head with a sigmoid activation outputs per-point confidence scores $\mathbf{c}_p, \mathbf{c}_q \in [0,1]^n$.
These confidence values are normalized into $\mathbf{w}_p = \mathbf{c}_p / \overline{\mathbf{c}_p},\ \mathbf{w}_q = \mathbf{c}_q / \overline{\mathbf{c}_q}$, and incorporated as target marginals in an optimal transport formulation, while the cosine similarity between geometric and semantic features forms the affinity kernel $\mathbf{K} \in \mathbb{R}^{n \times n}$. 
We apply the Sinkhorn algorithm~\cite{sinkhorn} to solve for the transport plan $\Pi \in [0,1]^{n \times n}$ and derive the row-stochastic correspondence matrices $\mathbf{M}_{pq}, \mathbf{M}_{qp} \in [0,1]^{n \times n}$. 
Using these correspondences as projection operators, we compute soft matches $\mathbf{M}_{pq}\mathbf{Q}$ and $\mathbf{M}_{qp}\mathbf{P}$ via convex combination, and estimate the rigid transformation using weighted SVD~\cite{Umeyama}, where normalized confidence weights 
$\mathbf{w}_p$ and $\mathbf{w}_q$ emphasize reliable correspondences.
After the coarse phase, the estimated pose $(\Tilde{\mathbf{R}}_{pq}, \Tilde{\mathbf{t}}_{pq})$ is applied to transform the query cloud, producing a coarse aligned version used in the fine phase to predict the final refined pose $(\mathbf{\hat{R}}_{pq} \in SO(3),\ \mathbf{\hat{t}}_{pq} \in \mathbb{R}^3)$. 
During inference, iterative refinement is applied by repeatedly transforming the query cloud using the estimated pose, further improving alignment accuracy.

We next introduce the optimal transport based correspondence formulation in Sec.~\ref{sec:correspondence}, followed by pose estimation in Sec.~\ref{sec:pose}, semantic priors in Sec.~\ref{sec:semantic}, and confidence learning in Sec.~\ref{sec:conf}.

\subsubsection{Optimal Correspondence}\label{sec:correspondence}

In our formulation, the confidence of each point represents its likelihood of finding a valid correspondence in the counterpart point cloud. 
Thus, correspondence estimation between the query and reference can be interpreted as transporting these confidence weights from one distribution to the other. 
This perspective naturally casts the problem as an optimal transport (OT) task, where the transport plan satisfies marginal constraints approximately consistent with the predicted confidences. 
Such a formulation ensures globally balanced soft correspondences and prevents over concentration on a few dominant keypoints.  

We first construct an affinity kernel that combines geometric and semantic similarities as
\begin{equation}
\mathbf{K}_{[i,j]} =
\exp \!\big(\tfrac{1}{\tau}\,
    \langle \mathbf{G}_{p[i]}, \mathbf{G}_{q[j]} \rangle_{\!\cos}\!\big)
\!\big(1 + \langle \mathbf{S}_{p[i]}, \mathbf{S}_{q[j]} \rangle_{\!\cos}\!\big)^{\lambda / \tau},
\end{equation}
where $\tau$ and $\lambda$ are temperature and semantic prior weight hyper-parameters. 
To handle cases where the two point clouds exhibit different confidence distributions, we normalize the predicted confidences by their means to form target marginals
$\mathbf{w}_p = \mathbf{c}_p / \overline{\mathbf{c}_p}$ and 
$\mathbf{w}_q = \mathbf{c}_q / \overline{\mathbf{c}_q}$,
ensuring that $\sum_i \mathbf{w}_{p[i]} = \sum_j \mathbf{w}_{q[j]} = n$. 
This normalization preserves global balance and allows sparse high-confidence regions to distribute their mass over denser ones.  
The transport plan $\mathbf{\Pi} = \mathcal{S}(\mathbf{K}, \mathbf{w}_p, \mathbf{w}_q) \in [0,1]^{n \times n}$ is obtained using the Sinkhorn algorithm~\cite{sinkhorn} (denotes with $\mathcal{S}$), where
$\sum_j \mathbf{\Pi}_{[i,j]} \!\approx\! \mathbf{w}_{p[i]}$ and 
$\sum_i \mathbf{\Pi}_{[i,j]} \!\approx\! \mathbf{w}_{q[j]}$. 
We then normalize $\mathbf{\Pi}$ row-wise to derive directional, row-stochastic correspondence matrices, such as
$
\mathbf{M}_{pq[i,j]} =
\frac{\mathbf{\Pi}_{[i,j]}}{\sum_k \mathbf{\Pi}_{[i,k]}},
$
and
$
\mathbf{M}_{qp[i,j]} =
\frac{\mathbf{\Pi}^\top_{[i,j]}}{\sum_k \mathbf{\Pi}^\top_{[i,k]}}.
$
Here, $\mathbf{M}_{pq}$ and $\mathbf{M}_{qp}$ act as soft projection operators, mapping each point in one cloud to a convex combination of points in the other.
Accordingly, we denote $\mathbf{M}_{pq}\mathbf{Q}$ as the corresponding points of $\mathbf{P}$ in the $\mathbf{Q}$ space, and $\mathbf{M}_{qp}\mathbf{P}$ vice versa.

To regularize the correspondences, we enforce a cycle consistency constraint inspired by CycleGAN~\cite{cycleGAN}. 
A point be projected to the opposite domain and back with $\mathbf{M}_{pq}$ and $\mathbf{M}_{qp}$ should approximately reconstruct its original position. 
The cycle consistency loss is defined as
\begin{equation}
\begin{split}
    \mathcal{L}_{\mathrm{cycl}} =
    & \frac{1}{n}\sum_{i=1}^n \mathbf{w}_{p[i]}\!
        \Big( 1 - \phi_{\mathrm{cycl}}
            (\mathbf{P}, \mathbf{P}^{\mathrm{rec}})_{[i]} \Big) \\
    & + \frac{1}{n}\sum_{j=1}^n \mathbf{w}_{q[j]}\!
        \Big( 1 - \phi_{\mathrm{cycl}}
            (\mathbf{Q}, \mathbf{Q}^{\mathrm{rec}})_{[j]} \Big),
\end{split}
\end{equation}
where $\mathbf{P}^{\mathrm{rec}} = \mathbf{M}_{pq}\mathbf{M}_{qp}\mathbf{P}$ and 
$\mathbf{Q}^{\mathrm{rec}} = \mathbf{M}_{qp}\mathbf{M}_{pq}\mathbf{Q}$. 
The kernel $\phi_{\mathrm{cycl}}$ measures geometric similarity via a Gaussian RBF:
\begin{equation}
    \phi_{\mathrm{cycl}}(\mathbf{X}, \mathbf{Y})_{[i]}
    = \exp\!\big(
        -\alpha_{\mathrm{g}} \|\mathbf{X}_{[i]} - \mathbf{Y}_{[i]}\|_2^2
    \big),
\end{equation}
where $\alpha_{\mathrm{g}}$ is a geometric scaling parameter and $\mathbf{X}, \mathbf{Y} \in \mathbb{R}^{n\times3}$ are point clouds.

\subsubsection{Pose Estimation}\label{sec:pose}

Once the correspondence matrices and point-wise confidences are obtained, we estimate the 6DoF rigid transformation between the query and reference using a confidence-weighted SVD, Umeyama algorithm~\cite{Umeyama}, also known as corresponding point alignment~\cite{pytorch3d}. 
This algorithm takes two point sets with associated correspondences and per-point weights as input, and outputs the optimal rotation and translation minimizing the weighted least squares error.

For the consistency, we concatenate both sides correspondences with the original point clouds, and perform joint optimization as
\begin{equation}
\mathbf{\hat{R}}_{pq},\ \mathbf{\hat{t}}_{pq}= 
\mathcal{U}\!\big(
    [\mathbf{P} \,|\, \mathbf{M}_{qp} \mathbf{P}],\ 
    [\mathbf{M}_{pq} \mathbf{Q} \,|\, \mathbf{Q}], \
    [\mathbf{w}_{p} | \mathbf{w}_{q}]
\big),
\end{equation}
where $\mathcal{U}(\cdot)$ denotes the Umeyama solver, and $[\mathbf{X}|\mathbf{Y}] \in \mathbb{R}^{2n\times3}$ represents concatenation along the point dimension. 
The inverse transformation is naturally given by $\mathbf{\hat{R}}_{qp} = \mathbf{\hat{R}}_{pq}^\top$ and $\mathbf{\hat{t}}_{qp} = - \mathbf{\hat{R}}_{pq}^\top \mathbf{\hat{t}}_{pq}$. 
The transformed point clouds are thus
$\mathbf{P}^{\mathrm{pred}} = \mathbf{P}\mathbf{\hat{R}}_{pq} + \mathbf{1}\mathbf{\hat{t}}_{pq}^\top$ and 
$\mathbf{Q}^{\mathrm{pred}} = \mathbf{Q}\mathbf{\hat{R}}_{qp} + \mathbf{1}\mathbf{\hat{t}}_{qp}^\top$.

The estimated poses are optimized using a confidence-weighted Chamfer loss that measures geometric alignment between the transformed and target point clouds, defined as
\begin{equation}
\begin{split}
    \mathcal{L}_{\mathrm{pose}} = 
    &\ \frac{1}{n}\!\sum_{i=1}^{n}\!\mathbf{w}_{p[i]}\!
        \Big(1-\phi_{\mathrm{pose}}(\mathbf{P}, \mathbf{Q}^{\mathrm{pred}})_{[i]}\!\Big) \\
    &+\frac{1}{n}\!\sum_{j=1}^{n}\!\mathbf{w}_{q[j]}\!
        \Big(1-\phi_{\mathrm{pose}}(\mathbf{Q}, \mathbf{P}^{\mathrm{pred}})_{[j]}\!\Big),
\end{split}
\end{equation}
where $\phi_{\mathrm{pose}}$ is a Gaussian RBF kernel based on the Chamfer distance~\cite{dcd}:
\begin{equation}
\phi_{\mathrm{pose}}(\mathbf{X}, \mathbf{Y})_{[i]} =
\exp\!\Big(
    -\alpha_{\mathrm{g}} \min_{j \in \{1,\dots,n\}}
    \|\mathbf{X}_{[i]}-\mathbf{Y}_{[j]}\|_2^2
\Big),
\end{equation}
with $\alpha_{\mathrm{g}}$ a geometric scaling hyperparameter and $\mathbf{X}, \mathbf{Y} \in \mathbb{R}^{n\times3}$ denoting point clouds.

\subsubsection{Semantic Priors}\label{sec:semantic}

Semantic features provide valuable cues for establishing reliable correspondences, as they help constrain matches within semantically consistent regions. 
To leverage such cues, we incorporate semantic priors derived from vision foundation models~(VFMs) such as DINOv2~\cite{dinov2}, which produce patch level RGB embeddings capturing high level visual semantics.
However, we observe that raw DINO features, while semantically rich, often encode mixed information unrelated to object parts, resulting in feature inconsistencies across viewpoints. 
Points belonging to the same semantic part may still exhibit noticeable differences in feature space, degrading cross-view matching quality. 
To improve their robustness, we adopt the self label refinement strategy of STEGO~\cite{stego}, which applies energy based clustering for semantic denoising and yields more stable and consistent features.
In our model, a lightweight semantic head (an MLP) projects the DINO features $\mathbf{F}_p$ and $\mathbf{F}_q$ into lower dimensional semantic embeddings $\mathbf{S}_p$ and $\mathbf{S}_q$. 
The semantic head is trained with the same loss as STEGO, promoting feature consistency across corresponding regions while filtering noise.

To softly constrain the correspondence using semantic similarity, we define a semantic consistency loss that penalizes correspondences assigned to semantically dissimilar points. 
Formally,
\begin{equation}
\begin{split}
    \mathcal{L}_{\mathrm{sem}} =
    & \frac{1}{n}\!\sum_{i=1}^n \mathbf{w}_{p[i]} 
        \Big(1 - \phi_{\mathrm{sem}}(\mathbf{M}_{pq}, \mathbf{S}_{p}, \mathbf{S}_{q})_{[i]} \Big) \\
    & + \frac{1}{n}\!\sum_{j=1}^n \mathbf{w}_{q[j]}
        \Big(1 - \phi_{\mathrm{sem}}(\mathbf{M}_{qp}, \mathbf{S}_{q}, \mathbf{S}_{p})_{[j]} \Big),
\end{split}
\end{equation}
where $\phi_{\mathrm{sem}}$ measures feature similarity through a Gaussian RBF kernel weighted by the correspondence matrix:
\begin{multline}
    \phi_{\mathrm{sem}}(\mathbf{M}_{uv}, \mathbf{U}, \mathbf{V})_{[i]} =
    \\
    \sum_{j=1}^{n} \mathbf{M}_{uv[i,j]}\,
    \exp\!\big(-\alpha_{\mathrm{f}} (1 - \langle\mathbf{U}_{[i]}, \mathbf{V}_{[j]} \rangle_{\!\cos})\big),
\end{multline}
where $\mathbf{M}_{uv} \in [0,1]^{n \times n}$ is a row-stochastic correspondence matrix, 
$\mathbf{U}, \mathbf{V} \in \mathbb{R}^{n \times d}$ are semantic feature matrices, 
and $\alpha_{\mathrm{f}}$ is a scaling hyperparameter. 
This loss encourages correspondences to align semantically coherent regions while maintaining the soft matching flexibility of OT.

\subsubsection{Confidence Learning}\label{sec:conf}

The absence of ground-truth confidence labels poses a key challenge for unsupervised training. 
We address this by constructing pseudo confidence labels derived from the model's own geometric and semantic consistency quality. 
Intuitively, a reliable (high confidence) point should incur low loss values, corresponding to high values of the Gaussian RBF kernels $\phi_{\mathrm{cycl}}$, $\phi_{\mathrm{pose}}$, and $\phi_{\mathrm{sem}}$. 
Hence, these kernel responses can be interpreted as soft inlier likelihoods that jointly reflect geometric, semantic, and cyclic consistency.

Formally, we define a composite pseudo likelihood as
$
\phi_{\mathrm{tot}}(\cdot)
= \phi_{\mathrm{cycl}}(\cdot)\,
  \phi_{\mathrm{pose}}(\cdot)\,
  \phi_{\mathrm{sem}}(\cdot),
$
to generate pseudo confidence labels:
$
\mathbf{z}_p = 
\phi_{\mathrm{tot}}(\mathbf{P}, \mathbf{P}^{\mathrm{rec}}, \mathbf{Q}^{\mathrm{pred}}, \mathbf{M}_{pq}, \mathbf{S}_p, \mathbf{S}_q),
$
and
$
\mathbf{z}_q = 
\phi_{\mathrm{tot}}(\mathbf{Q}, \mathbf{Q}^{\mathrm{rec}}, \mathbf{P}^{\mathrm{pred}}, \mathbf{M}_{qp}, \mathbf{S}_q, \mathbf{S}_p).
$
These labels provide graded supervision instead of binary inlier-outlier signals, enabling the confidence branch to learn to down weight uncertain points rather than discard them. 
By jointly fusing cues from geometric reconstruction, pose alignment, and semantic consistency, the pseudo labels capture complementary aspects of correspondence reliability and yield well calibrated confidence predictions.
The confidence is optimized using a binary cross entropy~($\mathrm{BCE}$) loss:
\begin{equation}
    \mathcal{L}_{\mathrm{conf}} =
    \mathrm{BCE}\big(\mathbf{c}_p,\, \mathrm{detach}(\mathbf{z}_p)\big)
    + \mathrm{BCE}\big(\mathbf{c}_q,\, \mathrm{detach}(\mathbf{z}_q)\big),
\end{equation}
where $\mathrm{detach}$ indicates a stop-gradient operation to prevent gradients from propagating into other loss terms, ensuring that $\mathcal{L}_{\mathrm{conf}}$ serves purely as supervision for confidence.

\section{Experiments}

\begin{table*}[th]
\centering
\resizebox{\linewidth}{!}{
\begin{tabular}{llllccccr}
\toprule
Method & Supervision & Modality & Reference & LM-O~\cite{lmo}~$\uparrow$ & TUD-L~\cite{tudl}~$\uparrow$ & YCB-V~\cite{ycbv}~$\uparrow$ & Mean~$\uparrow$ & Time (s)~$\downarrow$ \\
\cmidrule(lr){1-4} \cmidrule(lr){5-8} \cmidrule(lr){9-9}
PPF~\cite{ppf} & None & PC & Image & 29.7 & 14.8 & 38.3 & 27.6 & 11.8 \\
FPFH+MAC~\cite{fpfh, mac} & None & PC & Image & 22.5 & 22.1 & 49.6 & 31.4 & 136.9 \\
FPFH+RANSAC~\cite{fpfh, ransac} & None & PC & Image & 31.0 & 31.0 & 50.0 & 37.3 & 6.4 \\
PPF+ICP~\cite{ppf, besl1992method} & None & PC & Image & 44.7 & 29.1 & \underline{66.8} & 46.9 & 14.3 \\
Robust OT~\cite{robustot} & None & PC+RGB & Image & 45.5 & 66.3 & 66.0 & 59.3 & \textbf{4.0}\\
FreeZe~\cite{freeze} & None & PC+RGB & Image & 45.5 & \underline{68.3} & 65.5 & 59.8 & 53.0 \\
Dustbin OT~\cite{dustbin-ot} & None & PC+RGB & Image & \underline{50.2} & 67.6 & 65.4 & \underline{61.1} & 4.2\\
\scc{blue}{8}{\methodAbbr~(Unsupervised)} & \scc{blue}{8}{None} & \scc{blue}{8}{PC+RGB} & \scc{blue}{8}{Image} & \scc{blue}{8}{\textbf{56.7}} & \scc{blue}{8}{\textbf{73.8}} & \scc{blue}{8}{\textbf{75.9}} & \scc{blue}{8}{\textbf{68.8}} & \scc{blue}{8}{\textbf{4.0}}\\
\cmidrule(lr){1-9}
RPM-Net~\cite{rpmnet} & GT Pose & PC & Image & 38.9 & 28.0 & 23.8 & 30.2 & \textbf{3.2}\\
FCGF+MAC~\cite{fcgf, mac} & GT Pose & PC & Image & 33.9 & 48.3 & 51.0 & 44.4 & 60.5 \\
FCGF+RANSAC~\cite{fcgf, ransac} & GT Pose & PC & Image & 38.9 & 59.0 & 57.6 & 51.8 & 11.0 \\
GeDi~\cite{gedi} & GT Pose & PC & Image & 42.8 & 67.3 & 60.6 & 56.9 & 48.9 \\
SAM-6D~\cite{sam6d} & GT Pose & PC+RGB & Posed Image & 54.5 & 29.7 & 68.1 & 50.8 & 4.2 \\
UnoPose~\cite{UnoPose} & GT Pose & PC+RGB & Image & \underline{58.7} & \underline{71.0} & \textbf{83.1} & \underline{70.9} & \underline{3.7} \\
\scc{blue}{8}{\methodAbbr~(Supervised)} & \scc{blue}{8}{GT Pose} & \scc{blue}{8}{PC+RGB} & \scc{blue}{8}{Image} & \scc{blue}{8}{\textbf{60.8}} & \scc{blue}{8}{\textbf{80.0}} & \scc{blue}{8}{\underline{80.5}} & \scc{blue}{8}{\textbf{73.8}} & \scc{blue}{8}{4.0}\\
\bottomrule
\end{tabular}
}
\caption{
Quantitative comparison of single-reference novel object pose estimation methods on LM-O, TUD-L, and YCB-V. 
\methodAbbr\ achieves state-of-the-art performance in both supervised and unsupervised settings, with comparable inference speed.
}
\vspace{-0.5em}
\label{tab:result}
\end{table*}

\begin{figure*}[th]
    \centering
    \includegraphics[width=\linewidth]{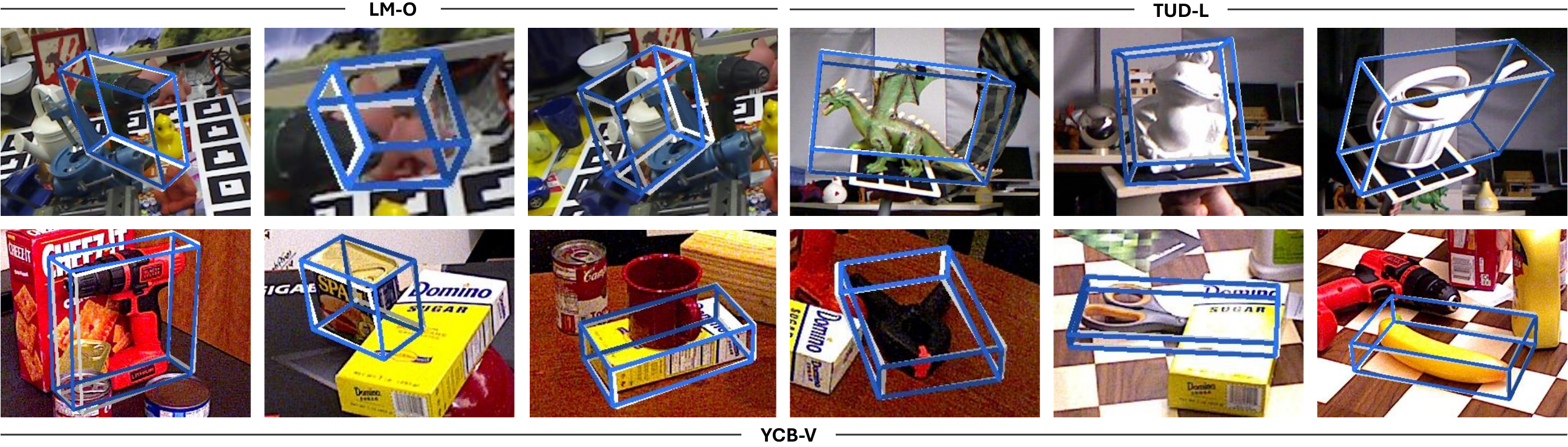}
    \caption{
    Qualitative results of unsupervised \methodAbbr\ on LM-O, TUD-L, and YCB-V datasets. 
    Blue bounding boxes represent the estimated poses, while white boxes denote ground-truth poses.
    }
    \vspace{-0.6em}
    \label{fig:result}
\end{figure*}

\subsection{Experimental Settings}

\paragraph{Datasets and Benchmarks.}
Following MegaPose~\cite{megapose}, we train \methodAbbr\ on two large-scale datasets: Google Scanned Objects~\cite{gso} and ShapeNet~\cite{shapenet}. 
Together, these datasets contain approximately $2{,}000{,}000$ RGB-D images of over $50{,}000$ objects covering diverse shapes and materials. 
For evaluation, we adopt BOP~\cite{bop} benchmarks, TUD-L, LM-O, and YCB-V~\cite{tudl, lmo, ycbv}, which test generalization to unseen objects and complex scenes. 
Specifically, TUD-L includes 600 images of 3 geometrically intricate objects; LM-O contains 200 images of 8 objects in a cluttered tabletop scene; and YCB-V provides 900 images of 21 household objects, sometimes heavily occluded. 

\paragraph{Implementation Details.}
We use the same backbone as our direct baseline UnoPose~\cite{UnoPose}: the coarse phase encoder-decoder is implemented as standard geometric transformer~\cite{geoTransformer}, while the fine phase adopts a sparse-to-dense variant introduced in \cite{sam6d}.
We also re-implement Robust OT~\cite{robustot} and Dustbin OT~\cite{dustbin-ot} with these backbones.
For fair comparison, we use the same query-reference pairing and the same segmentation masks as UnoPose for all models.
We train using the ADAM optimizer~\cite{adam} with an initial learning rate of $10^{-4}$.
The model is trained for 3 epochs with a batch size of 32.
The loss weights are set to $\gamma_{\mathrm{cycl}}=0.5$, $\gamma_{\mathrm{pose}}=1$, $\gamma_{\mathrm{sem}}=1$, and $\gamma_{\mathrm{conf}}=10$. 
Hyperparameters are $\lambda=3$, $\tau=0.01$, $\alpha_{\mathrm{f}}=4$, and $\alpha_{\mathrm{g}}=60$.
We randomly sample 1024 points from both query and reference point clouds as input to the fine phase, and 256 points via farthest point sampling~\cite{fps} for the coarse phase. 
Additionally, we train a supervised variant of \methodAbbr\ by only replacing the Chamfer distance in $\mathcal{L}_{\mathrm{pose}}$ with the point-wise distance between the predicted and ground-truth transformed points 
$\big(\mathbf{P}\mathbf{R}_{pq} + \mathbf{1}\mathbf{t}_{pq}^{\top}\big)_{[i]}$ 
and 
$\big(\mathbf{P}\mathbf{\hat{R}}_{pq} + \mathbf{1}\mathbf{\hat{t}}_{pq}^{\top}\big)_{[i]}$.

\paragraph{Evaluation Metrics.}
For pose evaluation, we follow the BOP protocol, report mean Average Precision~(mAP) under three standard error metrics: VSD, MSSD, and MSPD (see~\cite{bop} for definitions).
We also report the average inference time, including both segmentation and pose estimation modules. 
For overlapping evaluation, we report the Intersection-over-Union~(IoU) between the predicted and the GT overlapping regions.

\subsection{Results}

\subsubsection{Object Pose Estimation}

Quantitative and qualitative results for single-reference novel object pose estimation are presented in Tab.~\ref{tab:result} and Fig.~\ref{fig:result}, respectively. 
\methodAbbr, when trained in an unsupervised manner, not only outperforms all other unsupervised baselines but also outperforms most supervised methods. 
Compared with the state-of-the-art supervised approach UnoPose~\cite{UnoPose}, our unsupervised model achieves comparable performance, with only a $2.1\%$ gap on average. 
Notably, on benchmark with relatively complicated object shapes, TUD-L, our unsupervised method exceeds UnoPose for $2.8\%$. Meanwhile, on benchmarks characterized by cluttered scenes and heavily occlusions, LM-O and YCB-V, performance gap increases, suggesting that unsupervised training still has room for improvement in handling complicated environments.
When trained with pose supervision, \methodAbbr\ achieves the best overall performance across LM-O and TUD-L, outperforming all existing methods. 
In particular, the large improvement on TUD-L highlights the model's robustness in accurately capturing fine-grained geometric correspondences. Compared to other OT-based methods, Robust OT~\cite{robustot}, Dustbin OT~\cite{dustbin-ot}, and RPM-Net~\cite{rpmnet}, our method exceeds them on all benchmarks, indicating the efficiency of confidence marginal OT and network-predicted confidence rather than post hoc calibration.
Overall, these results demonstrate that \methodAbbr\ achieves performance competitive with supervised methods even without ground-truth supervision, and that its supervised variant further establishes a new state of the art. 
This validates the effectiveness of our confidence-aware OT formulation and the unsupervised network-predicted confidence.

\subsubsection{Overlapping Prediction}

\begin{figure*}[th]
    \centering
    \includegraphics[width=\linewidth]{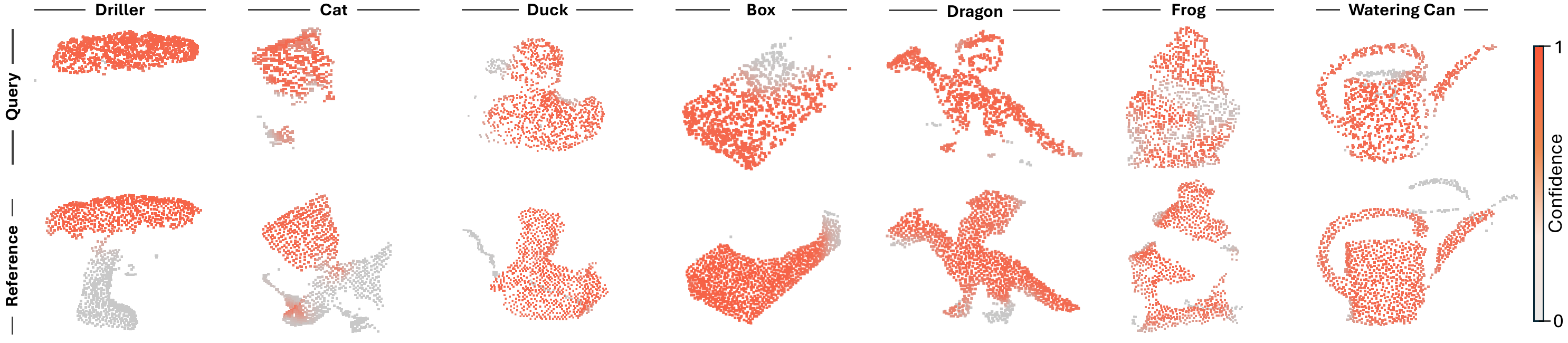}
    \caption{
    Visualization of predicted confidence from unsupervised \methodAbbr. 
    Our method effectively handles non-overlapping regions and outlier points by assigning low confidence to unreliable points.
    Poses are aligned for visualization.
    }
    \vspace{-0.5em}
    \label{fig:overlap}
\end{figure*}

\begin{table}[t]
\centering
\resizebox{1\linewidth}{!}{
\begin{tabular}{lcccc}
\toprule
Method & Dragon & Frog & Watering Can & Mean\\
\cmidrule(lr){1-1} \cmidrule(lr){2-5}
UnoPose~\cite{UnoPose} & 70.0 & \textbf{72.2} & 59.1 & 67.1  \\
\scc{blue}{8}{COG (Supervised)} & \scc{blue}{8}{\textbf{72.9}} & \scc{blue}{8}{\underline{68.3}} & \scc{blue}{8}{\textbf{83.9}} & \scc{blue}{8}{\textbf{75.0}} \\
\scc{blue}{8}{COG (Unsupervised)} & \scc{blue}{8}{\underline{71.2}} & \scc{blue}{8}{64.4} & \scc{blue}{8}{\underline{81.2}} & \scc{blue}{8}{\underline{72.3}} \\
\bottomrule
\end{tabular}
}
\caption{Overlapping IoU results on TUD-L benchmark's objects.}
\vspace{-0.6em}
\label{tab:overlap}
\end{table}

As an essential step toward accurate pose estimation, we also evaluate the overlapping prediction derived from the confidence values. 
Predicted confidence values greater than $0.5$ are treated as positives, while others are negatives. 
Tab.~\ref{tab:overlap} reports the IoU between predicted and ground-truth overlapping regions on the TUD-L dataset, and qualitative examples are shown in the last 3 columns of Fig.~\ref{fig:overlap}.
\methodAbbr\ effectively distinguishes overlapping from non-overlapping areas by assigning high confidence to valid points and low confidence to outliers and semantic or geometric inconsistent points. 
Even without supervision, our method exceeds the supervised UnoPose's performance in average, and the visualization demonstrate that the learned confidence is both interpretable and robust for outliers.

\subsection{Ablation Studies}

\subsubsection{Modules and Losses}

\begin{table}[t]
\centering
\resizebox{1\linewidth}{!}{
\begin{tabular}{cccccccc}
\toprule
 & Correspondence & $\mathcal{L}_{\mathrm{sem}}$ & $\mathcal{L}_{\mathrm{cycl}}$ & VSD~$\uparrow$ & MSSD~$\uparrow$ & MSPD~$\uparrow$ & Mean~$\uparrow$\\
\cmidrule(lr){1-4} \cmidrule(lr){5-8}
A1) & Argmax        &            &            & 71.3 & 78.4 & 64.4 & 71.4 \\
A2) & Argmax        &            & \checkmark & 72.1 & 78.5 & 64.2 & 71.6 \\
A3) & Argmax        & \checkmark &            & 68.3 & 79.1 & 64.0 & 70.5 \\
A4) & Argmax        & \checkmark & \checkmark & 73.4 & 80.0 & 65.9 & 73.1 \\
S1) & Softmax       &            &            & 71.9 & 78.6 & 64.0 & 71.5 \\
S2) & Softmax       &            & \checkmark & 72.1 & 78.5 & 64.1 & 71.6 \\
S3) & Softmax       & \checkmark &            & 68.6 & 78.3 & 63.5 & 70.1 \\
S4) & Softmax       & \checkmark & \checkmark & 73.4 & 79.8 & 65.8 & 73.0 \\
U1) & Uniform OT    &            &            & 74.6 & 80.2 & 67.8 & 74.2 \\
U2) & Uniform OT    &            & \checkmark & 75.0 & 80.0 & 67.0 & 74.0 \\
U3) & Uniform OT    & \checkmark &            & 74.1 & \underline{82.1} & 69.0 & 75.0 \\
U4) & Uniform OT    & \checkmark & \checkmark & \underline{75.8} & 81.5 & 68.4 & 75.2 \\
C1) & Confidence OT &            &            & 74.8 & 80.3 & 67.9 & 74.3 \\
C2) & Confidence OT &            & \checkmark & 75.5 & 80.3 & 67.4 & 74.4 \\
C3) & Confidence OT & \checkmark &            & 74.9 & \textbf{82.5} & \textbf{69.6} & \underline{75.6} \\
\scc{blue}{8}{C4)} & \scc{blue}{8}{Confidence OT} & \scc{blue}{8}{\checkmark} & \scc{blue}{8}{\checkmark} & \scc{blue}{8}{\textbf{76.5}} & \scc{blue}{8}{82.0} & \scc{blue}{8}{\underline{69.1}} & \scc{blue}{8}{\textbf{75.9}} \\
\bottomrule
\end{tabular}
}
\caption{Modules and losses ablation on YCB-V benchmark. Correspondence refers to whether use argmax, softmax, uniform marginal OT, or confidence marginal OT to estimate correspondence, $\mathcal{L}_{\mathrm{sem}}$ and $\mathcal{L}_{\mathrm{cycl}}$ refer to the use of these loss terms.}
\vspace{-0.5em}
\label{tab:ablation}
\end{table}

To evaluate the contribution of different correspondence formulations and loss terms, we conduct ablation studies on the YCB-V benchmark. 
As shown in Tab.~\ref{tab:ablation}, both uniform-marginal OT and our confidence-marginal OT outperform the argmax and softmax baselines by roughly $2\!-\!3\%$, demonstrating the advantage of globally balanced correspondences over discrete or row-normalized mappings. 
Among OT-based variants, the confidence-marginal formulation further improves the mean performance, validating the effectiveness of incorporating learned confidences as non-uniform marginals. 
And regarding the loss terms, adding $\mathcal{L}_{\mathrm{sem}}$ consistently improves the MSSD and MSPD metrics, while $\mathcal{L}_{\mathrm{cycl}}$ mainly benefits VSD, indicating that semantic consistency enhances geometric alignment, and cycle consistency improves visible-region correspondence. 
Overall, the combination of confidence marginal OT with both auxiliary losses yields the best mean performance.

\subsubsection{OT Parameters}

\begin{table}[t]
\centering
\resizebox{\linewidth}{!}{
\begin{tabular}{cccccc}
\toprule
 & Sinkhorn Iteration & Semantic Priors & mAP~$\uparrow$ & ENT~$\downarrow$  & $\Delta\mathrm{Marginal}$~$\downarrow$\\
\cmidrule(lr){1-3} \cmidrule(lr){4-6}
C4-1g) & 1 &            & 73.1 &  21.1 & 0.35\\
C4-2g) & 2 &            & 73.2 &  23.0 & 0.27\\
C4-4g) & 4 &            & 73.2 & 24.1 & \underline{0.21} \\
C4-8g) & 8 &            & 73.0 & 25.7 & \textbf{0.16} \\
C4-1s) & 1 & \checkmark & \underline{75.8} & \textbf{9.2} & 0.48 \\
\scc{blue}{8}{C4-2s)} & \scc{blue}{8}{2} & \scc{blue}{8}{\checkmark} & \scc{blue}{8}{\textbf{75.9}} & \scc{blue}{8}{\underline{10.5}} & \scc{blue}{8}{0.38}\\
C4-4s) & 4 & \checkmark & \underline{75.8} & 11.4 & 0.29 \\
C4-8s) & 8 & \checkmark & 75.7 & 12.0 & 0.23\\
\bottomrule
\end{tabular}
}
\caption{OT parameters ablation on YCB-V benchmark. Effective Number of Tokens~(ENT) reflects the number of points receiving attention, lower values indicate sharper focus. $\Delta\mathrm{Marginal}$ indicates the distance between target marginals and result marginals.}
\label{tab:ablation2}
\end{table}

We further analyze the influence of parameters in the OT formulation, as summarized in Tab.~\ref{tab:ablation2}. 
We use the Effective Number of Tokens~(ENT), defined as the exponential of correspondence entropy, to measure the concentration of the transport plan---lower values indicate sharper, more focused correspondences. 
We also report $\Delta\mathrm{Marginal}$, the mean distance between target and resulting marginals, which reflects the marginal alignment quality.
When semantic priors are injected into the affinity kernel, mAP improves notably while ENT decreases, indicating that correspondences become more compact and semantically coherent. 
This confirms that semantic guidance helps the OT solver focus on meaningful, consistent regions across views. 
In contrast, increasing the number of Sinkhorn iterations beyond two slightly reduces performance. 
Although additional iterations reduce $\Delta\mathrm{Marginal}$, they also make the correspondence map more diffuse, which weakens the geometric precision of the convex combination. 
Considering this trade-off, we adopt two iterations in our final setting to balance OT marginal accuracy and correspondence sharpness.

\subsubsection{Iterative Refinement}

\begin{figure}[t]
    \centering
    \includegraphics[width=1\linewidth]{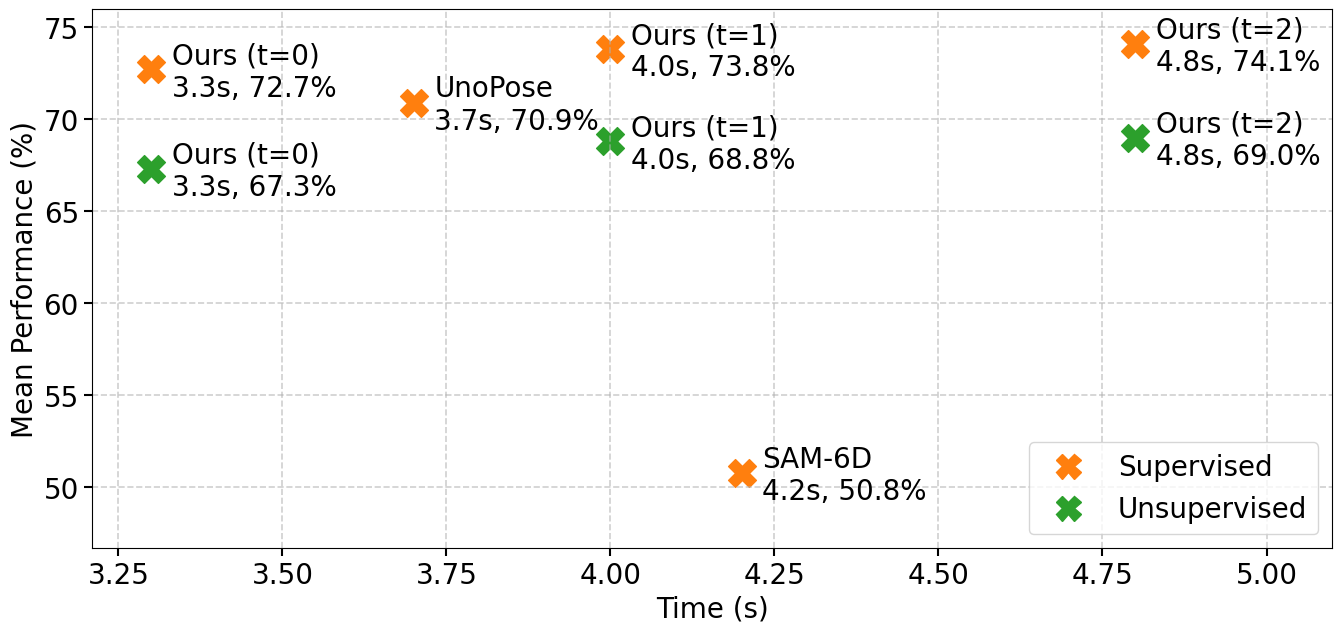}
    \caption{
    Performance vs runtime~(including segmentation, DINO feature extraction, and pose estimation). $t$ indicates the number of refinement iterations.
    }
    \label{fig:time}
\end{figure}

We further evaluate the performance gain and computational overhead introduced by iterative refinement on a single \texttt{NVIDIA RTX 3090} GPU. 
As shown in Fig.~\ref{fig:time}, increasing the number of refinement iterations consistently improves mean performance, though with diminishing returns. 
A single refinement already boosts both supervised and unsupervised \methodAbbr\ by over $1\%$, while the improvement becomes marginal beyond two iterations. 
Considering the trade-off between runtime and accuracy, we adopt one refinement iteration as our default setting.
\section{Conclusion}

We presented \methodName~(\methodAbbr), an unsupervised framework for single reference novel object pose estimation by solving a confidence-aware optimal transport problem. COG achieves balanced and robust correspondences that improve pose accuracy under occlusions by integrating point-wise confidence as transport marginals. And incorporating semantic priors from vision foundation models further enhances its semantic consistency. Furthermore, the confidence learned in an unsupervised manner from geometric, semantic, and cycle consistency enables \methodAbbr\ to down-weight unreliable regions without external labels.
Extensive experiments demonstrate that \methodAbbr\ achieves performance comparable to state-of-the-art supervised approaches, validating the effectiveness of our unsupervised formulation. 
Overall, COG provides a principled and scalable direction toward generalizable, unsupervised object pose estimation.
\newpage
{
    \small
    \bibliographystyle{ieeenat_fullname}
    \bibliography{ref}
}

\clearpage
\setcounter{page}{1}
\maketitlesupplementary
\appendix

\section*{Contents}
\begin{itemize}
    \item \textbf{Section~\ref{sec:model_detail}: Model Details}
        \begin{itemize}
            \item \ref{subsec:sinkhorn}: Sinkhorn and Affinity Kernel
            \item \ref{subsec:semantic}: Semantic Denoising
            \item \ref{subsec:pseudo}: Pseudo Confidence Labels
        \end{itemize}
    \item \textbf{Section~\ref{sec:experiment_detail}: Experiment Details}
        \begin{itemize}
            \item \ref{subsec:efficiency_experiment}: Data Efficiency Analysis
            \item \ref{subsec:confidence_experiment}: Confidence Analysis
            \item \ref{subsec:symmetric}: Symmetry Analysis
            \item \ref{subsec:addition_experiment}: Extended Results and Visualizations
        \end{itemize}
    \item \textbf{Section~\ref{sec:fail}: Limitations}
        \begin{itemize}
            \item \ref{subsec:segmentation}: Segmentation Failure
            \item \ref{subsec:unsupervised}: Unsupervised Limitation
        \end{itemize}
\end{itemize}

\section{Model Details}\label{sec:model_detail}

\subsection{Sinkhorn and Affinity Kernel}
\label{subsec:sinkhorn}

To compute differentiable cross-view correspondences under non-uniform confidence marginals, we adopt the entropy-regularized optimal transport (OT) framework with a log-domain Sinkhorn~\cite{sinkhorn} solver. Let $\mathbf{K}\in\mathbb{R}^{n\times n}$ denote the affinity matrix, and let $\mathbf{w}_{\mathrm{row}}=\mathbf{w}_p,\mathbf{w}_{\mathrm{col}}=\mathbf{w}_q$ be the target marginals obtained by normalizing point-wise confidences from the query and reference point clouds. The Sinkhorn algorithm iteratively updates the dual variables $\mathbf{u},\mathbf{v}\in\mathbb{R}^{n}$ and recovers a transport plan satisfying the prescribed marginals (Alg.~\ref{alg:sinkhorn}). The log-sum-exp implementation ensures numerical stability and prevents underflow in high-dimensional spaces.

For explanatory decomposition, we first define the OT cost as the negative similarity, and decompose the OT cost into a geometric similarity term and a nonlinear scaled semantic consistency term:
\begin{equation}
\mathbf{C}_{[i,j]}
=
- \langle \mathbf{G}_{p[i]}, \mathbf{G}_{q[j]} \rangle_{\cos}
- \lambda \log\!\big(1 + \langle \mathbf{S}_{p[i]}, \mathbf{S}_{q[j]} \rangle_{\cos}\big),
\label{eq:cost_decomp}
\end{equation}
where the semantic term is expressed in a logarithmic form. This design yields a semantic modulation that encourages semantically coherent matches without overwhelming the geometric similarity when semantic features are relatively similar. On the other hand, when the semantic similarity becomes negative, it yields a strong penalty for semantically inconsistent matches. This nonlinear shape prevents excessive amplification for already similar semantic pairs, while still sharply suppressing mismatched or unrelated regions. In practice, since cosine similarity may take negative values close to $-1$ and $\log(1+x)$ close to $-\infty$, a small stability constant $\epsilon=10^{-6}$ is added inside the logarithmic semantic term for stable training.
And the regularized OT problem is written as:
\begin{equation}
\min_{\mathbf{\Pi}\ge 0}
\ 
\langle \mathbf{C},\mathbf{\Pi}\rangle
\;+\;
\tau
\sum_{i,j}\mathbf{\Pi}_{[i,j]} \log \mathbf{\Pi}_{[i,j]},
\label{eq:reg_ot_rewrite}
\end{equation}
subject to the constraints
$
\mathbf{\Pi}\,\mathbf{1} = \mathbf{w}_{\mathrm{row}}
$
and
$
\mathbf{\Pi}^{\top}\mathbf{1} = \mathbf{w}_{\mathrm{col}}.
$
Under entropy regularization, the minimizer admits the Gibbs form 
$
\mathbf{\Pi}_{[i,j]} \propto \exp\!\left(-\tfrac{1}{\tau}\mathbf{C}_{[i,j]}\right).
$
Substituting Eq.~\eqref{eq:cost_decomp} yields the following factorized affinity kernel:
\begin{equation}
\begin{aligned}
    \mathbf{K}_{[i,j]} =& \exp\!\left(-\tfrac{1}{\tau}\mathbf{C}_{[i,j]}\right) \\
    =&
    \exp\!\left(
    \tfrac{1}{\tau}
    \langle \mathbf{G}_{p[i]},\mathbf{G}_{q[j]} \rangle_{\cos}
    \right)
    \big(
    1+\langle \mathbf{S}_{p[i]},\mathbf{S}_{q[j]} \rangle_{\cos}
    \big)^{\lambda/\tau}.
\end{aligned}
\label{eq:kernel_final_rewrite}
\end{equation}
which injects geometric affinity with semantic coherence, providing a soft semantic prior.

\begin{algorithm}[t]
\caption{Log-Domain Sinkhorn with Target Marginals}
\label{alg:sinkhorn}
\begin{algorithmic}[1]
\STATE \textbf{Input:} affinity $\mathbf{K}$, marginals $\mathbf{w}_{\mathrm{row}}, \mathbf{w}_{\mathrm{col}}$, iterations $T$
\STATE \textbf{Output:} transport matrix $\mathbf{\Pi}$
\STATE Initialize dual variables: $\mathbf{u}\!\gets\!\mathbf{0}^{n}$,\quad $\mathbf{v}\!\gets\!\mathbf{0}^{n}$
\FOR{$t = 1,\dots,T$}
    \STATE
    $\displaystyle
    \mathbf{u}_{[i]} \gets 
    \log \mathbf{w}_{\mathrm{row}[i]}
    - \log\!\sum_{j}\exp\!\left(\log \mathbf{K}_{[i,j]} + \mathbf{v}_{[j]}\right)
    $
    \vspace{-0.1em}
    \STATE
    $\displaystyle
    \mathbf{v}_{[j]} \gets 
    \log \mathbf{w}_{\mathrm{col}[j]}
    - \log\!\sum_{i}\exp\!\left(\log \mathbf{K}_{[i,j]} + \mathbf{u}_{[i]}\right)
    $
\ENDFOR
\STATE Recover transport plan:
\[
\mathbf{\Pi}_{[i,j]}
=
\exp\!\left(
\log\mathbf{K}_{[i,j]} + \mathbf{u}_{[i]} + \mathbf{v}_{[j]}
\right).
\]
\end{algorithmic}
\end{algorithm}


\subsection{Semantic Denoising}\label{subsec:semantic}

\begin{figure}[t]
    \centering
    \includegraphics[width=\linewidth]{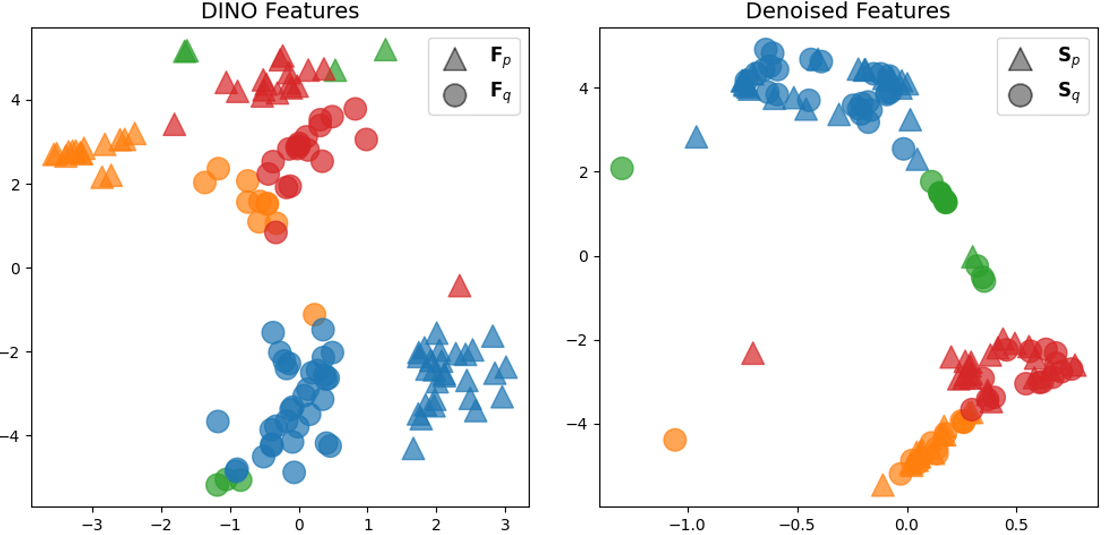}
    \caption{
    t-SNE results of DINO features on query and reference $\mathbf{F}_p, \mathbf{F}_q$, and denoised semantic features $\mathbf{S}_p, \mathbf{S}_q$, colored by k-means part segmentation. With semantic denoising, the distance between points from different views (triangle and circle) with the same semantic part is reducing dramatically. 
    }
    \label{fig:semantic}
\end{figure}

As we indicated in main script, we follow STEGO~\cite{stego} for semantic denoising. Fig.~\ref{fig:semantic} shows the t-SNE embeddings of the raw DINO~\cite{dinov2} features $F_p,F_q$ and the denoised semantic features $S_p,S_q$ from the query and reference views.
After semantic denoising, features belonging to the same semantic parts from different viewpoints become significantly closer in the embedding space, indicating improved cross-view semantic consistency.
This demonstrates that the lightweight semantic head effectively filters view-dependent noise in DINO features and provides more stable guidance for the subsequent OT correspondence.

\subsection{Pseudo Confidence Labels}\label{subsec:pseudo}

\begin{figure}[t!]
    \centering
    \includegraphics[width=\linewidth]{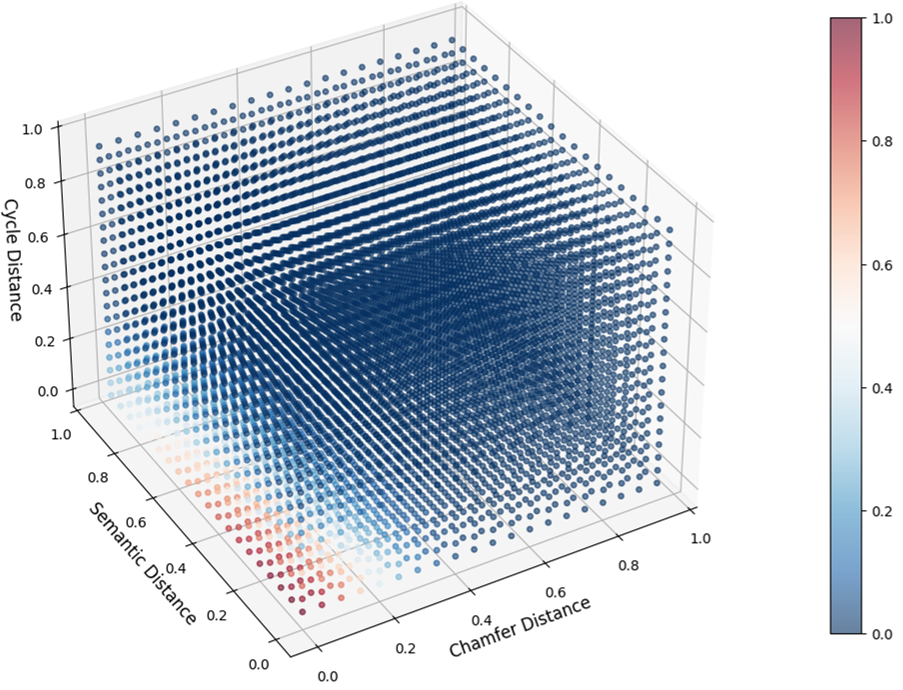}
    \caption{
    Visualization of pseudo confidence label generation. 
    Points with smaller Chamfer, cycle, and semantic distances receive higher confidence labels.
    }
    \vspace{-0.5em}
    \label{fig:pseudo}
\end{figure}

As an essential supervision signal of point validity, pseudo confidence labels are generated from Gaussian kernels defined over geometric ($\phi_{\mathrm{cycl}}, \phi_{\mathrm{pose}}$) and semantic ($\phi_{\mathrm{sem}}$) distances. 
As illustrated in Fig.~\ref{fig:pseudo}, points exhibiting small Chamfer, semantic, and cycle distances are assigned high confidence labels, while inconsistent or noisy points which contains larger distance receive low confidence labels. 

With such design, the training process can be interpreted as an EM-like process that unfolds across epochs rather than within a single optimization step. 
Specifically, the previous training iterations act as the \emph{E-step}, where the network estimates latent variables---soft correspondences and poses---and derives pseudo confidence labels based on the current model state. 
The current iteration then plays the role of the \emph{M-step}, in which these pseudo confidence labels supervise the confidence branch, guiding the model toward more accurate correspondence and pose estimation within high confidence points. 
Through this implicit alternating process across iterations, \methodAbbr\ progressively reinforces the consistency between confidence, correspondence, and pose, leading to stable performance without external supervision.

\paragraph{Stability.}
\textbf{Locality consistency}: Intuitively, once a point receives a high pseudo confidence label, its neighboring points, being geometrically close and semantically consistent, naturally inherit similar high confidence labels. This locality-aware consistency propagates stable supervision across nearby regions, making subsequent confidence predictions more coherent over iterations.
\textbf{Normalized confidence}: To further ensure robustness in early training (when confidence labels are close to zero), all losses except $\mathcal{L}_{\mathrm{conf}}$ are weighted by the normalized confidence $\mathbf{w}=\mathbf{c}/\overline{\mathbf{c}}$ rather than the raw confidence $\mathbf{c}$. Consequently, even if all predicted confidences start near zero, the optimization reduces to a uniform weighting scheme, preventing degenerate gradients and maintaining stable learning dynamics.
\textbf{Detached labels}: In addition, the pseudo confidence labels used in $\mathcal{L}_{\mathrm{conf}}$ are detached from backpropagation, avoiding harmful feedback loops that would push low confidence points farther.

In practice, with such design, the training process consistently behavior across all datasets, and we observe no signs of collapse or instability.

\section{Experiment Details} \label{sec:experiment_detail}

\subsection{Data Efficiency Analysis} \label{subsec:efficiency_experiment}

\begin{table}[t]
\centering
\resizebox{\linewidth}{!}{
\begin{tabular}{lcccc}
\toprule
Method & LM-O~\cite{lmo} & TUD-L~\cite{tudl} & YCB-V~\cite{ycbv} & Mean~$\uparrow$ \\
\cmidrule(lr){1-1} \cmidrule(lr){2-5}
DINO+PHS~\cite{dinov2}& 24.7 & 16.2 & 45.2 & 28.7 \\
COG (1\% data) & 52.9 & 66.9 & 73.6 & 64.5\\
COG (5\% data) & 55.7 & 70.5 & 75.7 & 67.3 \\
COG (25\% data) & 55.1 & 70.7 & 75.4 & 67.2\\
COG (50\% data) & 55.4 & 71.8 & 74.7 & 67.3 \\
COG (100\% data) & 56.7 & 73.8 & 75.9 & 68.8\\
\bottomrule
\end{tabular}
}
\caption{
Quantitative comparison of DINO with pose-hypothesis-scoring~(PHS) and \methodAbbr\ trained with different fractions of the training data.
}
\label{tab:result_scale}
\end{table}

As a task requires dataset-agnostic generalization, the data efficiency is important. To evaluate the data efficiency of our model, we train \methodAbbr\ with progressively smaller subsets of the training data. 
Specifically, we randomly sample $(1\%, 5\%, 25\%, 50\%)$ of the full training set (2M images) and train our unsupervised model on each subset. 
For comparison with training-free pure semantic features, we replace the OT module with one-to-one correspondence based on DINO~\cite{dinov2} feature similarity and apply a pose-hypothesis-scoring (PHS) scheme, which generates $6,000$ pose candidates and selects the one with the highest matching score, following the coarse pose selection strategy used in~\cite{sam6d, UnoPose}.

The results in Tab.~\ref{tab:result_scale} show that even with only $1\%$ of the data (about 20k images), our method achieves a substantial performance gain over the DINO baseline, highlighting the importance of geometric reasoning and confidence learning for accurate pose estimation. 
Interestingly, increasing the data volume beyond $5\%$ leads to only marginal improvements on LM-O~\cite{lmo} and YCB-V~\cite{ycbv} benchmarks, while performance on TUD-L~\cite{tudl} benchmark exhibits a more noticeable gain. 
This difference reflects the characteristics of each benchmark: TUD-L contains objects with complex shapes that benefit from high-precision geometric alignment, whereas LM-O and YCB-V include more texture-rich objects, where semantic priors already provide strong guidance. 

Overall, these findings demonstrate that unlike prior approaches relying on large-scale data, CAD models, or pose supervision, \methodAbbr\ generalizes effectively even under extremely limited data and unsupervised setting, highlighting the data efficiency of our unsupervised framework.

\subsection{Confidence Analysis} \label{subsec:confidence_experiment}

\begin{figure}[t]
    \centering
    \includegraphics[width=\linewidth]{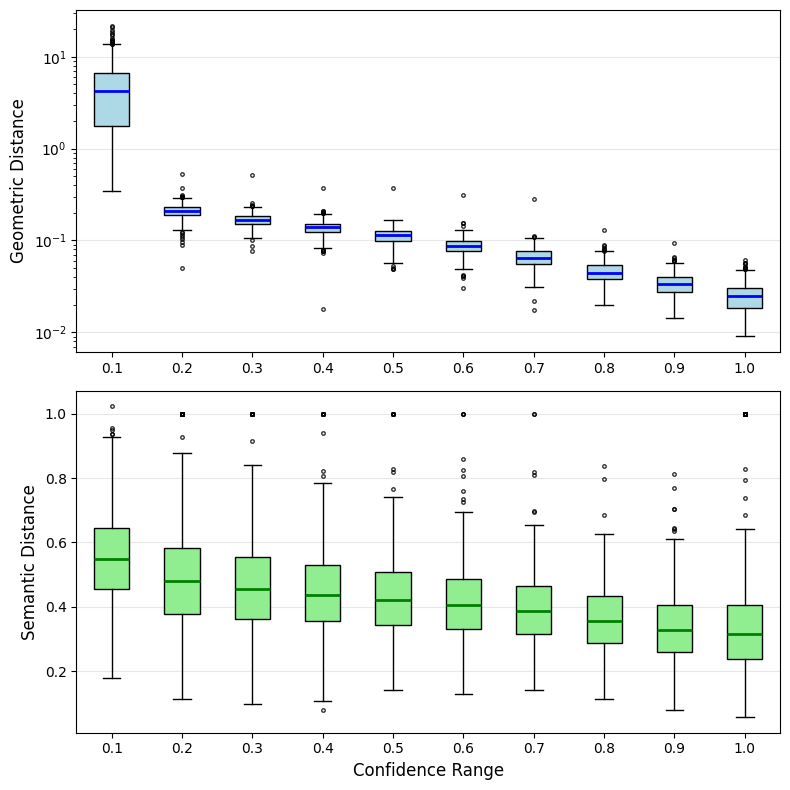}
    \caption{
    Relationship between predicted confidence and point-level distances for all test query-reference pairs in TUD-L~\cite{tudl} benchmark.
    The geometric distance (log scale) and semantic cosine distance are measured between each point and its nearest neighbor.
    }
    \label{fig:conf_pose}
\end{figure}

To further examine what the predicted confidence captures, 
we use all query-reference pairs in TUD-L~\cite{tudl}, and measure for each point its nearest corresponding point in the other view and compute two distances: geometric Euclidean distance in 3D space and semantic cosine distance ($1 - \langle \cdot, \cdot \rangle_{\mathrm{cos}}$) in feature space.

As shown in Fig.~\ref{fig:conf_pose}, both geometric and semantic distances show a consistent downward trend as confidence increases,
confirming that the confidence branch effectively learns to identify reliable correspondences, by jointly informed from geometric and semantic consistency.

\subsection{Symmetry Analysis} \label{subsec:symmetric}

\begin{figure}[t]
    \centering
    \includegraphics[width=\linewidth]{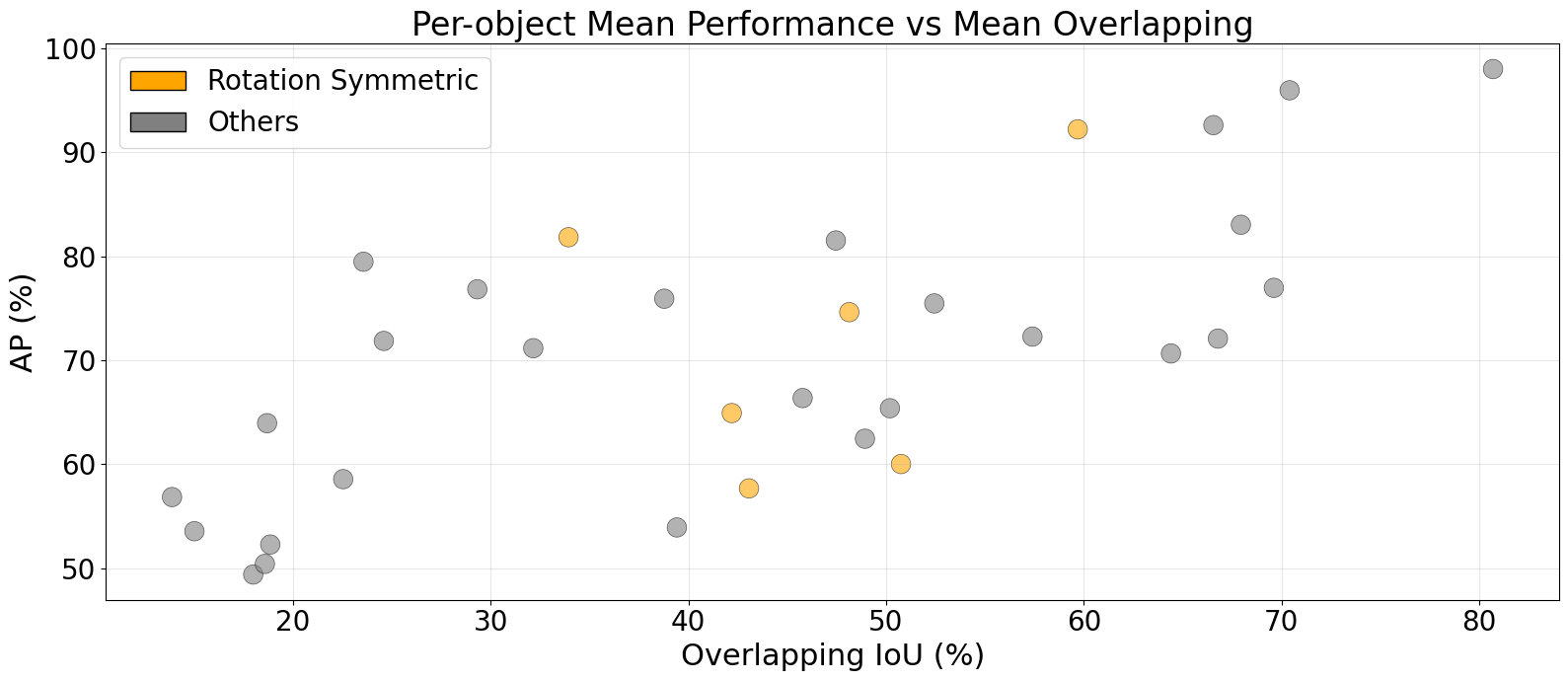}
    \caption{
    Per-object mean performance and overlapping.}
    \label{fig:symmetry}
\end{figure}

For rotationally symmetric objects, such as bottles and bowls, the geometric and semantic features of different points often become indistinguishable. This ambiguity can lead to multiple valid transport solutions and, consequently, inaccurate pose estimation.

To mitigate this risk, we incorporate positional embeddings into the fine-phase geometry encoder to sharpen diffused features. These coordinate-based embeddings provide a soft constraint, biasing the model toward selecting correspondences within a local neighborhood rather than distant, yet geometrically similar, points.
Furthermore, points with diffused correspondences typically fail the cycle consistency constraint, resulting in lower confidence scores.

As shown in Fig.~\ref{fig:symmetry}, our evaluation of objects with identical overlapping ratios demonstrates no significant performance degradation attributable to symmetry.

\subsection{Extended Results and Visualizations} \label{subsec:addition_experiment}

\begin{table}[t]
\centering
\resizebox{\linewidth}{!}{
\begin{tabular}{llcccc}
\toprule
Benchmark & Object & VSD~$\uparrow$ & MSSD~$\uparrow$ & MSPD~$\uparrow$ & Mean~$\uparrow$ \\
\cmidrule(lr){1-2} \cmidrule(lr){3-6}
\multirow{9}{*}{LM-O~\cite{lmo}}&ape          & 40.0 & 54.7 & 65.5 & 53.4 \\
&can          & 51.1 & 70.2 & 68.0 & 63.1 \\
&cat          & 37.6 & 53.1 & 58.1 & 49.6 \\
&driller      & 47.7 & 56.1 & 51.6 & 51.8 \\
&duck         & 53.1 & 59.2 & 68.6 & 60.3 \\
&eggbox       & 42.9 & 59.8 & 63.6 & 55.4 \\
&glue         & 38.7 & 55.2 & 54.8 & 49.6 \\
&holepuncher  & 59.5 & 69.5 & 71.7 & 66.9 \\
&\scc{blue}{8}{Average}       & \scc{blue}{8}{46.9} & \scc{blue}{8}{60.1} & \scc{blue}{8}{63.0} & \scc{blue}{8}{56.7} \\
\cmidrule(lr){1-6}

\multirow{4}{*}{TUD-L~\cite{tudl}}&dragon   & 64.2 & 83.3 & 81.0 & 76.2 \\
&frog     & 72.5 & 74.8 & 79.4 & 75.6 \\
&can      & 58.9 & 78.3 & 72.3 & 69.8 \\
&\scc{blue}{8}{Average}   & \scc{blue}{8}{65.2} & \scc{blue}{8}{78.8} & \scc{blue}{8}{77.6} & \scc{blue}{8}{73.8} \\
\cmidrule(lr){1-6}

\multirow{22}{*}{YCB-V~\cite{ycbv}}&master\_chef\_can    & 83.0 & 70.2 & 42.9 & 65.4 \\
&cracker\_box        & 71.0 & 79.6 & 62.5 & 71.0 \\
&sugar\_box          & 92.6 & 97.6 & 90.5 & 93.6 \\
&tomato\_soup\_can    & 84.9 & 75.5 & 64.8 & 75.0 \\
&mustard\_bottle     & 81.7 & 89.5 & 74.6 & 81.9 \\
&tuna\_fish\_can      & 72.8 & 56.1 & 53.8 & 60.9 \\
&pudding\_box        & 92.3 & 97.6 & 94.7 & 94.9 \\
&gelatin\_box        & 97.3 & 100.0 & 98.7 & 98.7 \\
&potted\_meat\_can    & 57.4 & 62.7 & 55.9 & 58.7 \\
&banana              & 75.7 & 87.7 & 64.7 & 76.0 \\
&pitcher\_base       & 87.4 & 86.5 & 59.9 & 77.9 \\
&bleach\_cleanser    & 80.1 & 88.2 & 68.0 & 78.8 \\
&bowl               & 87.1 & 97.3 & 91.1 & 91.8 \\
&mug                & 65.5 & 71.6 & 67.3 & 68.1 \\
&power\_drill        & 69.3 & 87.0 & 71.3 & 75.9 \\
&wood\_block         & 86.5 & 93.7 & 81.2 & 87.1 \\
&scissors           & 51.0 & 86.4 & 71.2 & 69.5 \\
&large\_marker       & 62.6 & 93.3 & 89.8 & 81.9 \\
&large\_clamp        & 62.0 & 87.5 & 79.7 & 76.4 \\
&extra\_large\_clamp  & 46.2 & 78.9 & 57.9 & 61.0 \\
&foam\_brick         & 80.8 & 86.4 & 83.3 & 83.5 \\
&\scc{blue}{8}{Average}   & \scc{blue}{8}{76.5} & \scc{blue}{8}{82.0} & \scc{blue}{8}{69.1} & \scc{blue}{8}{75.9} \\
\bottomrule
\end{tabular}
}
\caption{Per-object results on the LM-O~\cite{lmo}, TUD-L~\cite{tudl}, and YCB-V~\cite{ycbv} benchmarks.}
\label{tab:objects}
\end{table}

\begin{figure*}[t]
    \centering
    \includegraphics[width=0.98\linewidth]{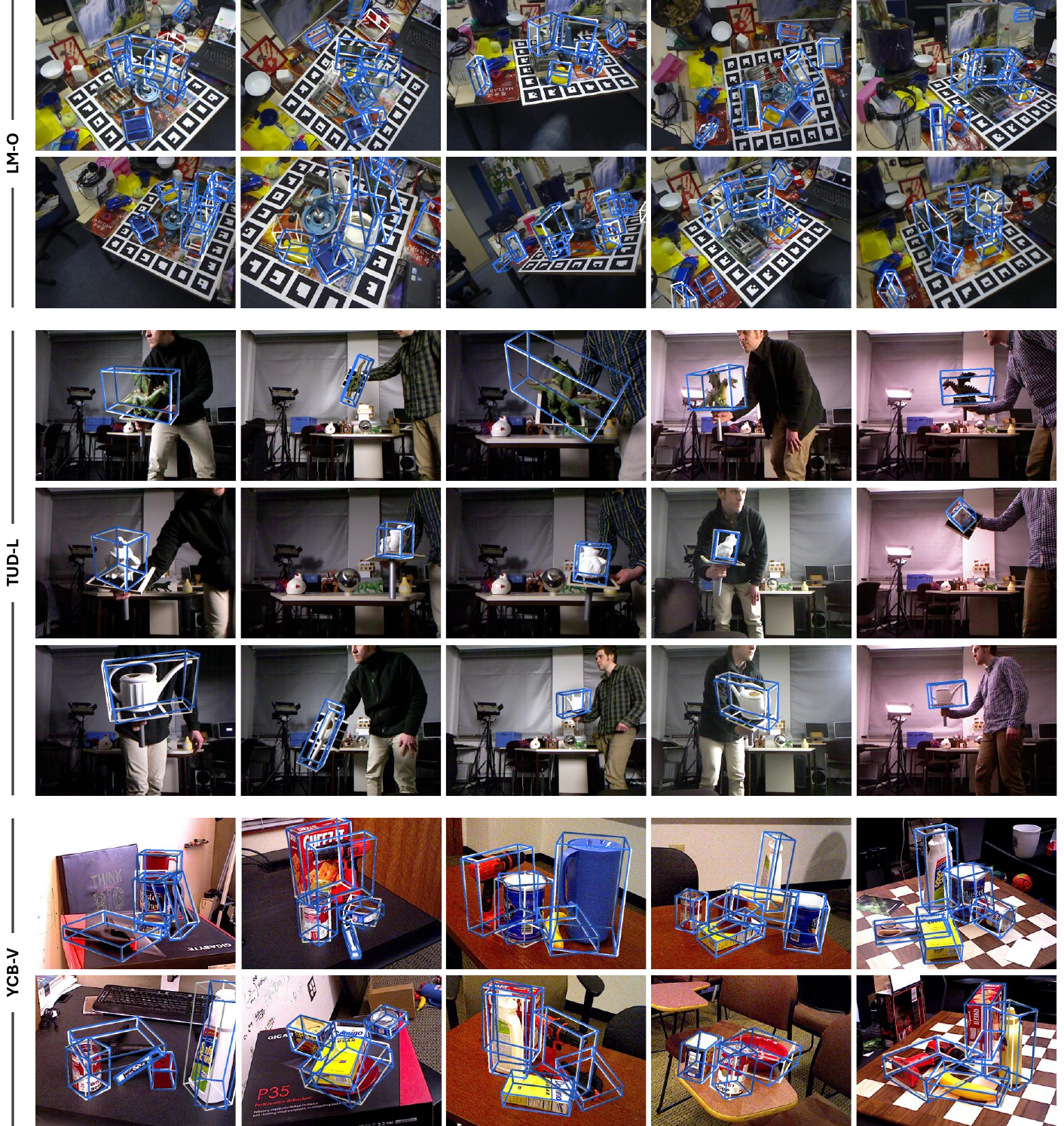}
    \caption{
    Results on 3 BOP benchmarks. Blue bounding boxes represent the estimated poses, while white boxes denote ground-truth poses.
    }
    \label{fig:supp_result}
\end{figure*}

In this section, we provide additional quantitative and qualitative results that complement the main experiments presented in the paper. 
Specifically, we report complete per-object results for all test objects on the LM-O, TUD-L, and YCB-V ~\cite{lmo, tudl, ycbv} benchmarks, as shown in Tab.~\ref{tab:objects}. 
We also include extended visualizations of estimated poses in Fig.~\ref{fig:supp_result}, illustrating the robustness of \methodAbbr\ under varying occlusion levels, viewpoint changes, and object shapes. 
These extended results further validate the conclusions drawn in the main paper, demonstrating that \methodAbbr\ achieves stable and accurate performance across diverse scenarios.

\section{Limitations} \label{sec:fail}

\begin{figure}[t]
    \centering
    \includegraphics[width=\linewidth]{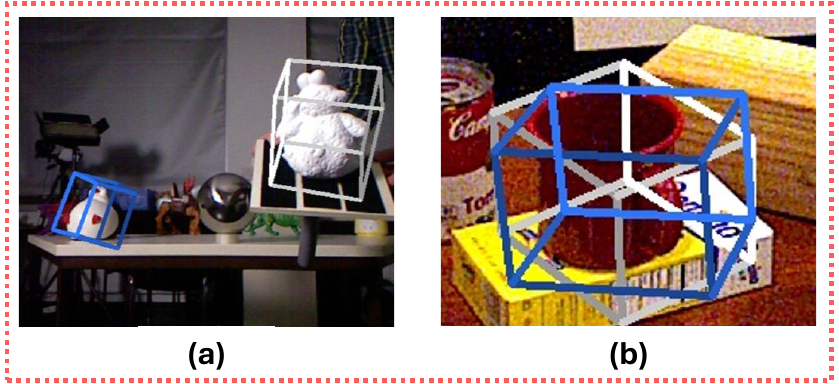}
    \caption{
    Failure cases of our method. (a) indicates the segmentation failure where segmentation mask is incorrect, and (b) indicates the unsupervised limitation discussed in Sec.~\ref{subsec:unsupervised}.
    }
    \label{fig:supp_fail}
\end{figure}

\subsection{Segmentation Failure} \label{subsec:segmentation}
Our framework relies on segmentation model to extract the masked depth for the point cloud inputs in the pre-processing. 
Therefore, segmentation errors can directly propagate to downstream pose estimation. 
As shown in Fig.~\ref{fig:supp_fail}~(a), if the segmentation model mistakenly includes different objects rather than the query object, the resulting point cloud may contain irrelevant geometry that cannot be recovered during pose estimation phase, leading to false correspondences and inaccurate.
In future work, integrating joint segmentation-pose optimization may mitigate such cascading errors. 

\subsection{Unsupervised Limitation} \label{subsec:unsupervised}
As an unsupervised framework, \methodAbbr\ optimizes pose implicitly, mainly through confidence-weighted Chamfer distance.
This objective encourages minimizing high confidence point distances rather than explicitly enforcing a globally correct pose transformation.
Consequently, when parts of the object contain sparse or noisy points, the network may prioritize aligning dense regions at the expense of small but semantically important parts. 
For example, as shown in Fig.~\ref{fig:supp_fail}~(b), when estimating the pose of a mug, the handle often contributes only few points; the network may achieve lower overall loss by ignoring this region and aligning only the main body of the mug. 
Although the introduction of semantic priors alleviates this issue to some extent, such priors remain soft constraints and cannot fully prevent these trade-offs.
Future work could explore hybrid supervision or structural regularization (\eg semantic part hard constraints) to better preserve critical fine-grained geometry during unsupervised optimization.

\end{document}